\pdfoutput=1

\documentclass[11pt]{article}

\usepackage{acl}

\usepackage{times}
\usepackage{latexsym}

\usepackage{color}

\usepackage[T1]{fontenc}

\usepackage[utf8]{inputenc}

\usepackage{microtype}

\usepackage{inconsolata}

\usepackage{graphicx}

\usepackage{booktabs}

\usepackage{amsmath}
\usepackage{amsfonts}
\usepackage{amssymb}

\title{Cross-Modal Robustness Transfer (CMRT): Training Robust Speech Translation Models Using Adversarial Text}

\author{Abderrahmane Issam \qquad {\bf Yusuf Can Semerci} \qquad {\bf Jan Scholtes} \qquad {\bf Gerasimos Spanakis} \\
        Department of Advanced Computing Sciences \\ 
        Maastricht University \\ 
        \small{\texttt{\{abderrahmane.issam, y.semerci, j.scholtes, jerry.spanakis\}@maastrichtuniversity.nl}}}
        
\begin{document}
\maketitle
\begin{abstract}
End-to-End Speech Translation (E2E-ST) has seen significant advancements, yet current models are primarily benchmarked on curated, "clean" datasets. This overlooks critical real-world challenges, such as morphological robustness to inflectional variations common in non-native or dialectal speech. In this work, we adapt a text-based adversarial attack targeting inflectional morphology to the speech domain and demonstrate that state-of-the-art E2E-ST models are highly vulnerable it. While adversarial training effectively mitigates such risks in text-based tasks, generating high-quality adversarial speech data remains computationally expensive and technically challenging. To address this, we propose Cross-Modal Robustness Transfer (CMRT), a framework that transfers adversarial robustness from the text modality to the speech modality. Our method eliminates the requirement for adversarial speech data during training. Extensive experiments across four language pairs demonstrate that CMRT improves adversarial robustness by an average of more than 3 BLEU points, establishing a new baseline for robust E2E-ST without the overhead of generating adversarial speech.
\end{abstract}

\section{Introduction}
In recent years, models that perform speech translation in an end-to-end fashion (E2E-ST)---mapping speech input directly to target-language text---have achieved remarkable results,  \cite{Dong_Wang_Zhou_Xu_Xu_Li_2021, tang-etal-2021-improving, ye-etal-2022-cross, fang-etal-2022-stemm, ouyang-etal-2023-waco, zhou-etal-2023-cmot, communication2023seamlessmultilingualexpressivestreaming, zhang2024-salign, zhang-etal-2025-representation}, often surpassing cascaded solutions where Automatic Speech Recognition (ASR) and Machine Translation (MT) models are combined \cite{anastasopoulos-etal-2021-findings, ye-etal-2022-cross, fang-etal-2022-stemm, ouyang-etal-2023-waco}. Unfortunately, the training and evaluation of E2E-ST models is mainly limited to datasets such as MuST-C \cite{di-gangi-etal-2019-must} and CoVoST 2 \cite{wang2020covost}, which fall short from capturing the full complexity of non-native and dialectal speech especially during spontaneous conversations. 

\begin{figure}[t!]
    \centering
    \includegraphics[width=0.48\textwidth]{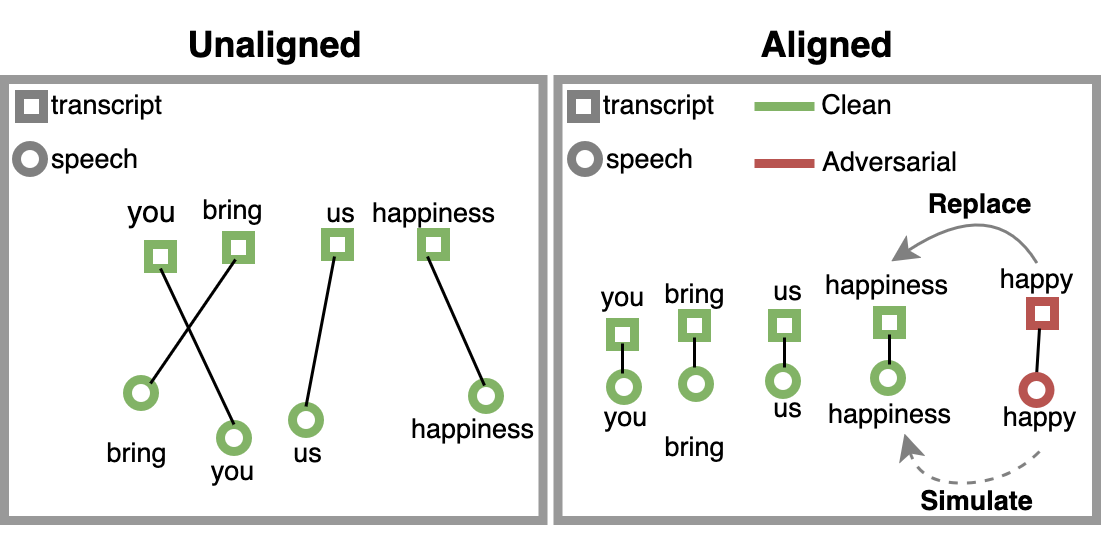}
    \caption[]{CMRT aligns speech and text semantic spaces (right). To \textbf{simulate} adversarial speech inflections, clean speech embeddings (e.g., "happiness") are \textbf{replaced} with adversarial text embeddings (e.g., "happy") during robustness fine-tuning.\footnotemark}
    \label{fig:cmrt_intro}
\end{figure}

\footnotetext{\scriptsize The example \textit{"you bring us happy"} is taken from \cite{Eaves_2011}.}
Non-native speech is characterized by distinct linguistic phenomena, including phonetic interference, disfluencies, and non-standard inflectional morphology \cite{moyer2011, gut2009non, prevost2000}. While datasets like CoVoST 2 include non-native speakers, these recordings consist of "read speech", which fails to capture the spontaneous morphological errors and fluency gaps typical of natural L2 production \cite{akande2013nonstandard, adejare2019acquisition}. In Neural Machine Translation (NMT), this gap is often bridged by fine-tuning on synthetic adversarial data \cite{anastasopoulos-etal-2019-neural, tan-etal-2020-morphin}; however, such advancements primarily benefit cascaded pipelines, leaving E2E-ST systems vulnerable. We attribute this neglect to the prohibitive cost of generating synthetic adversarial speech compared to text. While one could theoretically synthesize adversarial speech via Text-to-Speech (TTS), this approach assumes the availability of high-fidelity TTS for the source language and risks introducing acoustic artifacts that may degrade the E2E-ST model during fine-tuning. Moreover, the computational overhead of generating large-scale adversarial audio remains a significant barrier.

To overcome these limitations, we introduce Cross-Modal Robustness Transfer (CMRT), a framework that equips E2E-ST models with morphological robustness using only adversarial text data. Our approach is predicated on the hypothesis that if speech and text representations are sufficiently aligned in a shared semantic space, the model's sensitivity to perturbations in one modality can be addressed using the other. Specifically, by aligning word-level speech and text embeddings, we can simulate adversarial speech perturbations by injecting adversarial text embeddings directly into the speech manifold. CMRT operates in two stages: (1) Semantic Alignment, where we train an E2E-ST model to map speech and text into a shared representation space, building on previous work in E2E-ST \cite{fang-etal-2022-stemm, ouyang-etal-2023-waco}; and (2) Robustness Transfer, where we fine-tune the model by substituting segments of the speech representation with their adversarial text-based counterparts. As illustrated in Figure \ref{fig:cmrt_intro}, this allows the model to learn invariance to inflectional variations without requiring any adversarial speech.

We evaluate the effectiveness of CMRT in mitigating adversarial inflectional perturbations across four diverse language pairs: English to German (En-De), Catalan (En-Ca), and Arabic (En-Ar), as well as French to English (Fr-En). These directions are particularly relevant given that the number of non-native (L2) speakers for English and French exceeds native (L1) speakers by more than twofold \cite{eberhard2025}. To benchmark robustness, we adapt the MORPHEUS framework \cite{tan-etal-2020-morphin} from text to speech, generating a challenging adversarial test set characterized by inflectional variations. Our experimental results demonstrate that CMRT yields an average improvement of more than 3 BLEU points on adversarial speech. Notably, our method achieves performance competitive with models trained directly on synthetic adversarial speech, while demonstrating superior generalization on the original, clean CoVoST 2 test set---effectively overcoming the common "robustness-accuracy" trade-off.

To address the challenge of morphological robustness in speech translation, we make the following contributions:
\noindent
\begin{itemize}
    \item We build on bridging the modality gap works in E2E-ST \cite{fang-etal-2022-stemm,ouyang-etal-2023-waco} to train a model with a strongly aligned semantic space between speech and text.
    \item We introduce Speech-MORPHEUS, an adaptation of the MORPHEUS adversarial technique to the speech domain. We utilize this tool to demonstrate that state-of-the-art E2E-ST models remain highly susceptible to inflectional variations.
    \item We propose a novel fine-tuning paradigm that transfers robustness from text to speech. We show that injecting adversarial text perturbations into a shared latent space significantly improves E2E-ST performance on non-standard input without the need for adversarial speech.
\end{itemize}

\section{Related Works}

\textbf{Bridging the Modality Gap in E2E-ST} aims at alleviating the discrepancy between speech and text modalities in E2E-ST, which has been shown to be a crucial ingredient for improving the performance of E2E-ST models and closing the gap with cascaded models \cite{Liu2020BridgingTM, wang2020-bridging}. Various techniques were shown to be effective for this purpose \cite{yuchen2019, inaguma-etal-2021-source, ye-etal-2022-cross, fang-etal-2022-stemm, ouyang-etal-2023-waco, zhou-etal-2023-cmot, zhang-etal-2025-representation}. Among these techniques, \cite{ye-etal-2022-cross, ouyang-etal-2023-waco} directly optimize for this property by using contrastive learning to maximize the similarity between speech and text representations. \citet{fang-etal-2022-stemm} point to the fact that the translation model is still taking either speech or text as input, which limits cross-model transfer, therefore, they introduce mixup training \cite{zhang2018mixup}, where they train on a mixup of speech and text representations. In this work, we show that combining both contrastive learning and mixup training leads to the best alignment between speech and text representation spaces, and we leverage this for robustness fine-tuning.

\noindent
\textbf{Robustness in ST} previous work on robustness in speech translation has been limited to cascaded systems, where robustness has been studied either on ASR \cite{fukuda2018data, rosenberg2019-robust, Ravanelli2020-robust, chen2022-robust, prabhu-etal-2023-accented} or NMT \cite{li2018improvingrobustnessspeechtranslation,  martucci2021-robust, padfield-cherry-2021-inverted, wang-etal-2022-adversarially, indurthi-etal-2023-clad} side, while E2E-ST models robustness has been neglected especially robustness to inflectional variation. Fortunately, in the context of NMT, robustness to inflectional variation has been explored in multiple works that target robustness to non-native speech \cite{anastasopoulos-etal-2019-neural, tan-etal-2020-morphin, jayanthi-pratapa-2021-study}. Our work is inspired by MORPHEUS \cite{tan-etal-2020-morphin}, which introduces inflectional perturbations into clean sentences. We extend MORPHEUS to speech, and show that it significantly affects models trained on datasets such as CoVoST 2. Furthermore, we introduce a method for improving robustness to inflectional variation in E2E-ST using only adversarial text.

\section{Methodology}
\begin{figure}[t!]
    \centering
    \includegraphics[width=0.49\textwidth]{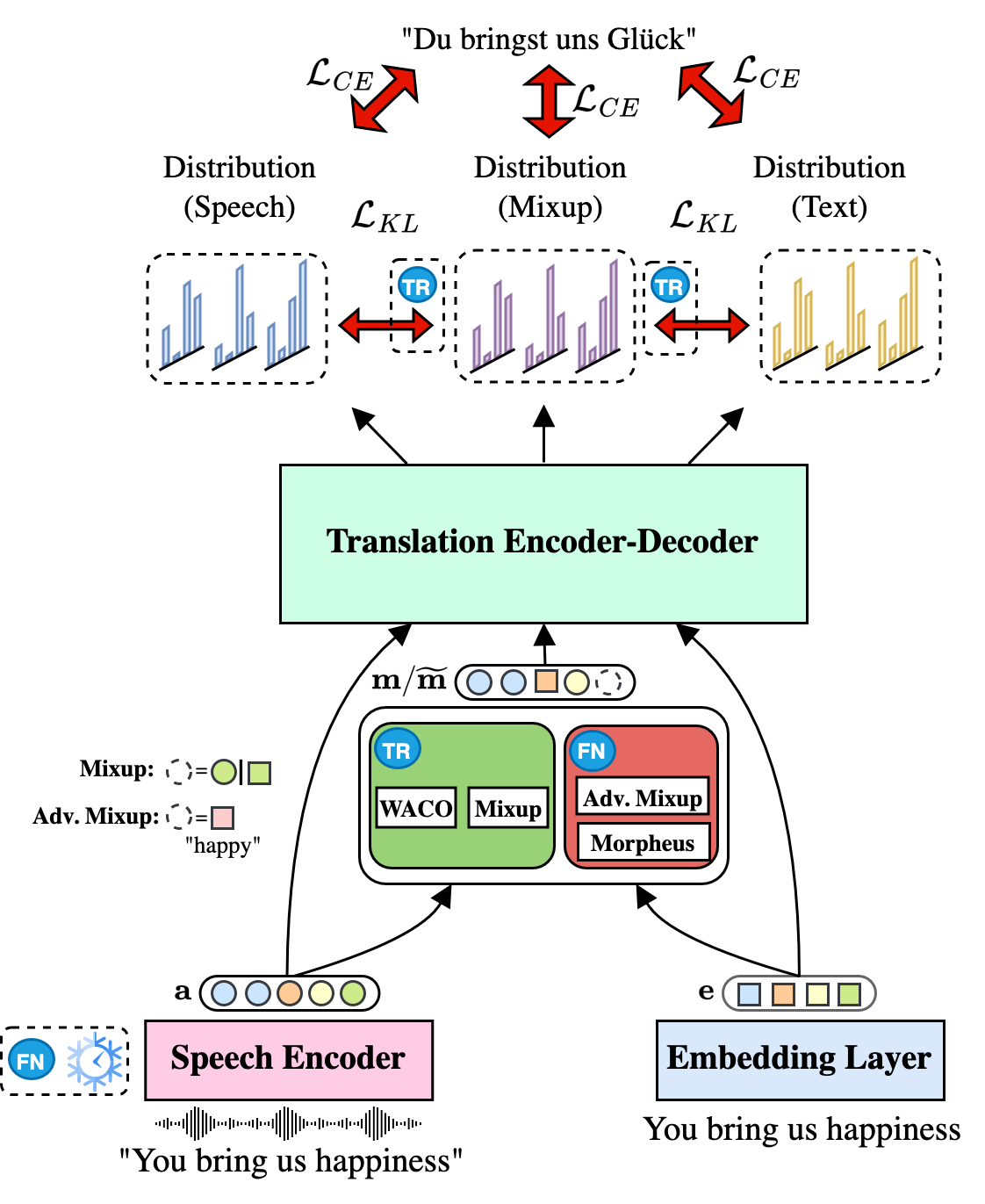}
    \caption{An overview illustration of our proposed method. Since our method is composed of two steps, CMRT-TR (\S\ref{sec:bridging_gap} and \S\ref{sec:cmrt_training}) followed by CMRT-FN (\S\ref{sec:cmrt_fn}), we refer to them as TR and FN respectively in the figure.}
    \label{fig:cmrt_method}
\vspace{-0.4cm}
\end{figure}

We begin this section by outlining the model architecture (\S\ref{sec:arch}). Then, in \S\ref{sec:cmrt} we describe our method, which is composed of two stages: a training stage described in \S\ref{sec:bridging_gap} and \S\ref{sec:cmrt_training}, and a fine-tuning stage described in \S\ref{sec:cmrt_fn}. Finally, in \S\ref{sec:speech_morpeus} we describe our method for extending MORPHEUS to speech. Figure \ref{fig:cmrt_method} illustrates the overview of our method.

Throughout this work, we denote the speech translation corpus as $D={(s,x,y)}$, where $s$ is the raw audio input, $x$ is the transcription, and $y$ is the translation in the target language. The triplets will be used during training, but at inference time the model will take the speech $s$ as input and generate the target translation $y$.

\subsection{Architecture}
\label{sec:arch}
Similarly to previous work in E2E-ST \cite{ye-etal-2022-cross, fang-etal-2022-stemm}, our model (Figure \ref{fig:cmrt_method}) is composed of a \textbf{speech encoder} that takes raw speech frames and transforms them into contextual embeddings, a text \textbf{embedding layer} that maps tokens into embeddings, and a \textbf{translation encoder-decoder} that takes either the text or speech embeddings and generates the target translation. 

\subsection{Cross-Modal Robustness Transfer (CMRT)}
\label{sec:cmrt}
CMRT consists of two main steps: first, it maximizes the alignment between speech and text semantic spaces (\S\ref{sec:bridging_gap}, \S\ref{sec:cmrt_training}), and second, it leverages the aligned speech and text embeddings for adversarial fine-tuning using only adversarial text embeddings (\S\ref{sec:cmrt_fn}).
\subsubsection{Aligning Speech and Text Representations}
\label{sec:bridging_gap}
Our hypothesis is that a strong alignment between speech and text representations is required for robustness to transfer from text to the speech modality, this is motivated by previous work showing that modality alignment allows for cross-modal transfer \cite{Castrejn2016LearningAC}. Furthermore, we believe that our method can be understood through modal complementarity \cite{wang2023deep}, where previous work has shown that speech models are weaker than text models at capturing syntactic features \cite{shen2023wave}, and by aligning speech and text representations our method complements the speech modality with textual syntactic features, which previous work has shown to be a crucial component of the robustness of MT models \cite{issam-etal-2025-representation}. Therefore, inspired by previous work on bridging the modality gap between speech and text modalities, we train the speech encoder model to output representations that are aligned with the text embedding representations. For this purpose, we use Word-Aligned Contrastive Learning (WACO) \cite{ouyang-etal-2023-waco}, where we train the model with contrastive learning to maximize the similarity between the speech and text embeddings of words in the sentence. However, this is not enough as the transformer encoder-decoder still takes either speech or text embeddings as input, which firstly, limits the cross-modal transfer between the modalities, and secondly for our purpose, the translation encoder-decoder has to be accustomed to receiving a mix of speech and text representations. Therefore, we rely on mixup training \cite{fang-etal-2022-stemm}, where we sample either speech or text embeddings of words based on a mixup probability. 

Both methods require word-level forced alignment between speech and text to identify words. For a sequence of words $w=[w_1,w_2,...,w_{|w|}]$, a sequence of speech embeddings $a=[a_1,a_2,...,a_{|a|}]$, and a sequence of text embeddings $e=[e_1,e_2,...,e_{|e|}]$, an alignment of the word $w_i$ consists of the start position $l^s_i$ and end position $r^s_i$ in the speech sequence, and the start position $l^t_i$ and end position $r^t_i$ in the text sequence. Therefore, the speech embeddings of the word $w_i$ are $a[l^s_i:r^s_i]$ while its text embeddings are $e[l^t_i:r^t_i]$. 

For WACO, we derive word level representations by mean pooling the speech and text embeddings of each word as follows:

{\small
\begin{align}
    f^s_i &= \text{MeanPool}(a[l^s_i:r^s_i]) \\
    f^t_i &= \text{MeanPool}(e[l^t_i:r^t_i])
\end{align}
}
$f^s_i$ and $f^t_i$ are considered as positive pairs, while other words in the batch are considered negatives. Therefore, the contrastive loss is computed as follows:

{\small 
\begin{equation} 
\ell_{\text{CTR}}(\mathcal{B}) = - \mathbb{E}_{i \in \mathcal{B}} \left[ \log \frac{\exp(\text{sim}(f_i^s, f_i^t) / \tau)}{\sum_{j \in \mathcal{B}} \exp(\text{sim}(f_i^s, f_j^t) / \tau)} \right] 
    \label{ctr_loss}
\end{equation}}
where $\mathcal{B}$ is the current batch, $\tau$ is the temperature, and sim is the cosine similarity function.

During mixup training, we similarly leverage alignments. For each word $w_i$ in the sequence, we choose either its speech embeddings or text embeddings based on a mixup probability $p^*$:

{\small
\begin{equation}
    \begin{aligned}
        m_i = 
        \begin{cases}
            a[l^s_i:r^s_i] & p < p^* \\
            e[l^t_i:r^t_i] & p \ge p^*
        \end{cases}
    \end{aligned}
\end{equation}
}

Where $p$ is sampled from a uniform distribution $\mathcal{U}(0, 1)$.

The final mixup sequence is the result of concatenating all $m_i$ segments:
{\small
\begin{equation}
m = Concat(m1,m2,...,m_{|w|})
\end{equation}
}

\subsubsection{Training}
\label{sec:cmrt_training}
Similar to previous work in E2E-ST, we start by pre-training the text embedding layer and the translation encoder-decoder on transcription-translation pairs $D={(x,y)}$ for the task of machine translation using cross-entropy as follows:

{\small
\begin{equation}
    \mathcal{L}_{MT} = -\mathbb{E}_{x,y} \log P(y|x)
    \label{mt_loss}
\end{equation}
}

The second involves multi-task learning with the addition of WACO and mixup training as introduced earlier. Below we briefly introduce each training objective. 

\noindent
\textbf{ST Loss: } We train with an ST objective where the speech encoder takes raw speech input $s$ and generates the target translation $y$. The ST loss is formalized as follows:

\begin{equation}
\small
    \mathcal{L}_{ST} = -\mathbb{E}_{s,x,y} \log P(y|s)
    \label{st_loss}
\end{equation}
\textbf{MT Loss: } To maximize cross modal transfer, we train with an MT objective as well. The MT loss is formalized as follows:
\begin{equation}
\small
    \mathcal{L}_{MT} = -\mathbb{E}_{s,x,y} \log P(y|x)
\end{equation}

\noindent
\textbf{CTR Loss: } We use contrastive learning as formalized in Equation \ref{ctr_loss}. The contrastive loss over the full dataset is formalized as:
\begin{equation}
\small
    \mathcal{L}_{CTR} = -\mathbb{E}_{\mathcal{B} \subseteq D_{s,x,y}}[\ell_{\text{CTR}}(\mathcal{B})]
\end{equation}

\noindent
\textbf{Mixup: } We train with cross-entropy where the translation encoder-decoder is supposed to take the mixup representations $m$ and translate to the target language:

\begin{equation}
\small
    \mathcal{L}_{M} = -\mathbb{E}_{m,y} \log P(y|m)
\end{equation}
This is an important step for familiarizing the model with translating from mixed representations. Another crucial step for bridging the modality gap is training with symmetric Kullback-Leibler (KL) divergence between the output probability distribution of the mixup $m$ and both the output probability distributions of the ST and MT tasks as follows:

{\small
\begin{equation}
    \begin{aligned}
         \mathcal{L}_{KL_{m \leftrightarrow s}} = \mathbb{D}_{KL}(P(y|s)||P(y|m)) + \\
         \mathbb{D}_{KL}(P(y|m)||P(y|s)) 
    \end{aligned}
\end{equation}
}

{\small
\begin{equation}
\begin{aligned}
        \mathcal{L}_{KL_{m \leftrightarrow x}} = \mathbb{D}_{KL}(P(y|x)||P(y|m)) + \\
        \mathbb{D}_{KL}(P(y|m)||P(y|x)) 
    \end{aligned}
\end{equation}
}

Therefore, the final loss combines the previous losses:
{\small
\begin{multline}
    \mathcal{L} = \mathcal{L}_{ST} + \mathcal{L}_{MT} + \lambda_{ctr}.\mathcal{L}_{CTR} + \mathcal{L}_{M} \\ + 
    \lambda_{kl}.(\mathcal{L}_{KL_{m \leftrightarrow s}}+ \mathcal{L}_{KL_{m \leftrightarrow x}})/2
\end{multline}
}
where $\lambda_{ctr}$ is the weight hyper-parameter for the contrastive loss, and $\lambda_{kl}$ is the weight for KL divergence loss that is part of the mixup training.

We term this stage of training \textbf{CMRT-TR}, where we train the model for speech translation with the priority of maximizing speech-text semantic alignment, which is achieved by combining both WACO and mixup training.
\subsubsection{Robustness Fine-tuning}
\label{sec:cmrt_fn}
This section builds on CMRT-TR to fine-tune the model for robustness to adversarial speech. In a typical adversarial fine-tuning setup, we would need adversarial speech examples $\widetilde{s}$ and their target translations $y$. However, generating synthetic text examples is much easier than generating speech examples, therefore, our method, which we term \textbf{CMRT-FN}, leverages the adversarial text examples $\widetilde{x}$ instead, along with clean speech $s$. We denote $\widetilde{D}={(s,\widetilde{x},y)}$ the adversarial dataset for training an adversarially robust model using CMRT-FN. 

Text adversarial attacks generally do not alter all words in the sentence but only a subset, otherwise, the meaning of the sentence is difficult to preserve. We denote the word indices of the words that were affected by the adversarial attack as $\widetilde{\mathcal{I}}$. For example, for a given sequence $w=[w_1, w_2, w_3, ...,w_{|w|}]$, a replacement adversarial attack that affects $\widetilde{\mathcal{I}}=[2,3]$ would replace the words $w_2$ and $w_3$ with adversarial copies $\widetilde{w_2}$ and $\widetilde{w_3}$ respectively.

Our strategy for fine-tuning CMRT-TR model is to introduce the adversarial text embeddings directly into the aligned speech manifold coming from the speech encoder, therefore, we freeze the speech encoder to maintain its alignment, and we train the rest of the model. Furthermore, we modify the mixup training strategy to use the adversarial embeddings of the inflection $\widetilde{w_i}$ if $i\in\widetilde{\mathcal{I}}$, otherwise, we apply standard mixup between clean speech and text embeddings. We denote the adversarial text embeddings of the word $\widetilde{w_i}$ as $\tilde{e}[\tilde{l}^t_i:\tilde{r}^t_i]$, where $\tilde{l}$ and $\tilde{r}$ are the start and end positions in the sequence of adversarial embeddings $\tilde{e}$. We term this step adversarial mixup:

\begin{equation}
\small
    \begin{aligned}
        \widetilde{m_i} = 
        \begin{cases}
            \tilde{e}[\tilde{l}^t_i:\tilde{r}^t_i] & i\in\widetilde{\mathcal{I}} \\
            a[l^s_i:r^s_i] & i\notin\widetilde{\mathcal{I}}, p < p^* \\
            e[l^t_i:r^t_i] & i\notin\widetilde{\mathcal{I}}, p \ge p^*
        \end{cases}
    \end{aligned}
\end{equation}

We use cross-entropy to train on adversarial mixup embeddings:
\begin{equation}
\small
    \mathcal{L}_{\widetilde{M}} = -\mathbb{E}_{\widetilde{m},y} \log P(y|\widetilde{m})
\end{equation}
Finally, we use asymmetric KL divergence to match output probability distribution of adversarial mixup with both the output probability distributions of ST and MT:

{\small
\begin{align}
    \mathcal{L}_{KL_{\widetilde{m} \rightarrow s}} = \mathbb{D}_{KL}(P(y|s)||P(y|\widetilde{m})) \\
    \mathcal{L}_{KL_{\widetilde{m} \rightarrow x}} = \mathbb{D}_{KL}(P(y|x)||(P(y|\widetilde{m})) 
\end{align}
}
Note that unlike CMRT-TR where we train with symmetric KL, we use asymmetric KL because the mixup representations are adversarial, therefore, we seek to train the model to handle adversarial noise by making the distribution of adversarial mixup similar to clean input. This part resembles virtual adversarial training (VAT) \cite{miyato2017adversarial}. We term this stage of training CMRT-FN, where we fine-tune the model to be robust to adversarial attacks. The final loss is formalized as:

\vspace{-0.8cm}
{\small
\begin{multline}
    \mathcal{L} = \mathcal{L}_{ST} + \mathcal{L}_{MT} + \mathcal{L}_{\widetilde{M}} \\ + 
    \lambda_{kl}.(\mathcal{L}_{KL_{\widetilde{m} \rightarrow s}}+ \mathcal{L}_{KL_{\widetilde{m} \rightarrow x}})/2
    \label{eq:cmrt_fn}
\end{multline}
}
\subsection{Speech-MORPHEUS}
\label{sec:speech_morpeus}
MORPHEUS \cite{tan-etal-2020-morphin} was introduced to simulate inflectional errors that are common in non-native or dialectal speech. It was confirmed through a human evaluation study that it generates sentences that are meaning-preserving, and that it is effective at simulating non-native speakers writing. Although this has been demonstrated in written language, inflectional errors can be more common in spoken language \cite{WEISSBERG200037, GARDNER2021104250}. This motivated our effort to introduce Speech-MORPHEUS, which simply extends MORPHEUS to speech. Similarly to MORPHEUS, we focus on verbs, nouns and adjectives in the sentence, and we find all candidate inflections with a similar Part-Of-Speech (POS) tag. We greedily search over the candidate inflections to find the ones that lead to significantly altering the model output. Given a sequence of words $w=[w_1,w_2, ...,w_{|w|}]$, MORPHEUS replaces each candidate word $w_i$ with each of its possible inflections until finding the set of candidate inflections $\widetilde{w}$. 

However, this process is infeasible in the context of ST where the input is continuous, therefore, we target the translation encoder-decoder instead by inflecting the transcriptions, then, we transform them into speech using a TTS model. However, even though this would work on languages such as English, it does not guarantee similar effect on languages with a high number of homophones such as French, where an inflection that might hurt the model in written form would not affect it in spoken form because the word is pronounced similarly to the original word. To address this, we add another step of filtering out candidate inflections that have the same phonemes as the original word. 

\section{Experiments}
\subsection{Dataset}
We conduct our experiments on CoVoST 2 dataset \cite{wang2020covost}, a large multilingual ST dataset that is based on Common Voice project \cite{ardila-etal-2020-common}. CoVoST 2 covers translation from 21 source languages to English and from English to 15 target languages. However, we only experiment with 4 language directions, namely En-De, En-Ca, En-Ar, Fr-En. We report the results on the test set. 

For adversarial data experiments, we use Speech-MORPHEUS (described in \S\ref{sec:speech_morpeus} and Appendix \ref{sec:exp_setup_details}) to generate an adversarial copy of 50k examples sampled from the train set, and a copy of the dev and test sets. We report the results on the test set. 

\subsection{Experimental Setup}
\noindent
\textbf{Preprocessing}
We use raw 16 bit 16kHz mono-channel speech input. We filter out examples with a number of speech frames that exceeds 480k or less than 1k. For text, we remove punctuation from the input transcriptions. We use uni-gram SentencePiece model with a vocabulary of 10k that is shared between the source and target.

Since our method requires word level alignments between speech and transcrption, we use NeMo Forced Aligner (NFA) \footnote{\scriptsize\texttt{ \sloppy\url{https://github.com/NVIDIA/NeMo/tree/main/tools/nemo\_forced\_aligner}}} which was shown to achieve state-of-the-art results in terms of alignment accuracy \cite{rastorgueva2023nemo}. 

\noindent
\textbf{Model}
Our model consists of two main components, a speech encoder which is composed of a pre-trained HuBERT model \cite{hsu-2021-hubert} in the case of En-X direction, and mHuBERT \cite{zanon-boito2024mhubert} in the case of Fr-En, plus two additional 1-dimensional convolution layers of kernel size 5, stride 2, padding 2, and hidden dimension 1024, to help shrink down the output representation over the time axis. The second component is a translation encoder-decoder model (pre-trained on CoVoST 2 train set), which has 6 encoder and 6 decoder layers. Each layer is comprised of 512 hidden units, 8 attention heads, and 2048 feed-forward hidden units. 

\noindent
\textbf{Baselines} \\
\textbf{\textit{MT-Transformer:} } The translation encoder-decoder with the embedding layer trained with the MT objective (Equation \ref{mt_loss}). \\
\textbf{\textit{HuBERT-Transformer: }} 
Similar architecture to our model trained with ST objective (Equation \ref{st_loss}). \\
\textbf{\textit{HuBERT-CMOT: }} 
Trained using state-of-the-art Cross-modal Mixup via Optimal Transport (CMOT) method \cite{zhou-etal-2023-cmot}. \\
\textbf{\textit{CMRT-TR: }}
Trained using WACO and Mixup as described in \S\ref{sec:cmrt_training}. \\
\textbf{\textit{TTS-Morpheus-FN: }} HuBERT-Transformer baseline fine-tuned on Speech-MORPHEUS synthetic train set of 50k examples with an ST objective. \\
\textbf{\textit{CMRT-FN: }}
CMRT-TR baseline fine-tuned with adversarial mixup training as described in \S\ref{sec:cmrt_fn}.

We provide further details on our experimental setup in Appendix \ref{sec:exp_setup_details}.
\section{Results and Analysis}

\subsection{Robustness Results}
\label{sec:cmrt_results}
\begin{table*}[ht]
    \centering
    \scriptsize
    \setlength{\tabcolsep}{4pt}
    \begin{tabular}{llccccccccccc}
        \toprule
        & \multicolumn{3}{c}{\textbf{En-De}} & \multicolumn{3}{c}{\textbf{En-Ca}} & \multicolumn{3}{c}{\textbf{En-Ar}} & \multicolumn{3}{c}{\textbf{Fr-En}} \\
        \cmidrule(lr){2-4} \cmidrule(lr){5-7} \cmidrule(lr){8-10} \cmidrule(lr){11-13}
         & Original & Morpheus & Clean & Original & Morpheus & Clean & Original & Morpheus & Clean & Original & Morpheus & Clean \\ 
        \midrule
        MT-Transformer & - & 13.2 & 26.7 & - & 20.2 & 34.6 & - & 11.3 & 18.3 & - & 21.3 & 32.3 \\
        \midrule
        HuBERT-Transformer & 21.4 & 14.4 & 25.3 & 27.4 & 19.7 & 32.7 & 15.7 & 12.8 & 19.2 & 28.4 & 20.5 & 27.9 \\
        HuBERT-CMOT & \textbf{21.8} & 14.6 & \textbf{25.9} & \textbf{28.2} & 18.4 & \textbf{33.5} & \textbf{16.2} & 14.5 & \textbf{19.5} & \textbf{30.9} & 22.0 & \textbf{30.9} \\
        CMRT-TR & 20.8 & 14.4 & 25.1 & 26.5 & 19.8 & 31.6 & 14.8 & 12.7 & 18.3 & 29.5 & 21.9 & 29.9 \\ 
        TTS-Morpheus-FN (50K) & 18.2 & \textbf{19.9} & 22.5 & 22.9 & \textbf{26.1} & 28.7 & 13.0 & \textbf{15.4} & 17.0 & 24.7 & 25.1 & 25.5 \\ 
        CMRT-FN (50K) & 19.9 & 17.4 & 24.0 & 25.4 & \underline{23.6} & 31.0 & 13.9 & 14.5 & 17.3 & 28.3 & 24.6 & 28.6 \\ 
        CMRT-FN & 20.3 & \underline{17.6} & 24.5 & 25.8 & \underline{23.6} & 31.7 & 14.4 & \underline{14.7} & 17.9 & 28.8 & \textbf{\underline{25.2}} & 29.2 \\
        \bottomrule
    \end{tabular}
    \caption{Results of CMRT-FN against baselines on 4 languages from CoVoST 2 dataset. Original represents the original CoVoST 2 test set, Morpheus shows the results on Speech-MORPHEUS adversarial copy of the same test set, and Clean shows the results on TTS of clean transcriptions. We underline the model scores that achieved the best result on Morpheus without using adversarial speech. On average, CMRT-FN improves robustness to MORPHEUS adversarial attack by more than 3 BLEU points compared to HuBERT-Transformer.}
    \label{tab:cmrt_results}
\end{table*}
Table \ref{tab:cmrt_results} shows that CMRT-FN is effective in transferring robustness from text to speech modality. Our method improves robustness by an average of 3.4 BLEU points on the 4 directions compared to HuBERT-Transformer without using any adversarial speech data (and 3.1 compared to CMRT-TR). CMRT-FN still falls short compared to fine-tuning on adversarial speech (an average of 1.4 BLEU points), as shown in the results of TTS-Morpheus-FN, which was fine-tuned on 50k examples from the train set that were adversarially attacked using Speech-MORPHEUS \footnote{\scriptsize We do not fine-tune on the full dataset because generating examples using the TTS model is time and resource consuming. We also consider 50k examples to be a reasonable amount of data for fine-tuning.}. However, this also leads to a significant drop in performance on the Original test set (i.e. 3.6 from HuBERT-Transformer), while our method witnesses a smaller drop in performance (i.e. 0.6 from CMRT-TR as the base model). Surprisingly, our method leads to a slightly better robustness on Fr-En, although this is partly due to CMRT-TR's performance on the Original data, it might suggest that our method can close the gap with adversarial fine-tuning on other source languages. 

To demonstrate that the drop in performance is due to Speech-MORPHEUS attack and not due to the quality of TTS data, Table \ref{tab:cmrt_results} also shows the results of applying the same TTS model on clean transcriptions, where the performance is even higher than the Original test set because the TTS speech is normalized. The table also shows the performance of the MT-Transformer on clean and adversarial MORPHEUS transcriptions, which we find to be correlated with the effect of Speech-MORPHEUS on HuBERT-Transformer.

CMOT achieves the best performance on Original and Clean test sets, however, when looking at adversarial robustness, this method still witnesses a significant drop in performance (7.4 on average). This demonstrates the fact that even state-of-the-art methods are vulnerable to adversarial attacks when trained on clean data, and our method can be advantageous in terms of robustness. In Appendix \ref{sec:cmrt_nllb_results} we take this further by fine-tuning a pre-trained state-of-the-art MT model, the results show that although this significantly improves the results, the model still witnesses a drop in performance on Morpheus (an average of 5.2), and CMRT leads to a significant improvement in robustness over the baseline (an average of 4.4) without using any adversarial speech data.

We note that although our method requires two stages of training, the second stage CMRT-FN is only a fine-tuning stage for 1/8th of the training stage \footnote{\scriptsize{{A CMRT-TR job on one H100 took 15:56:03 while CMRT-FN took only 1:46:19. This is significantly better than running the TTS model on GeForce RTX 2080 Ti which took 17:53:01 on En-De train sample of 50k examples.}}}, therefore, training with CMRT does not double the training time. Moreover, given the benefits of using CMRT, which alleviates the need for a TTS model as well the associated computational costs on a large dataset, the fine-tuning stage is a beneficial and necessary trade-off.

\subsection{Effect of Shared Semantic Space}
\label{cmrt_space}

\begin{figure}[ht]
    \centering
    \includegraphics[width=0.48\textwidth]{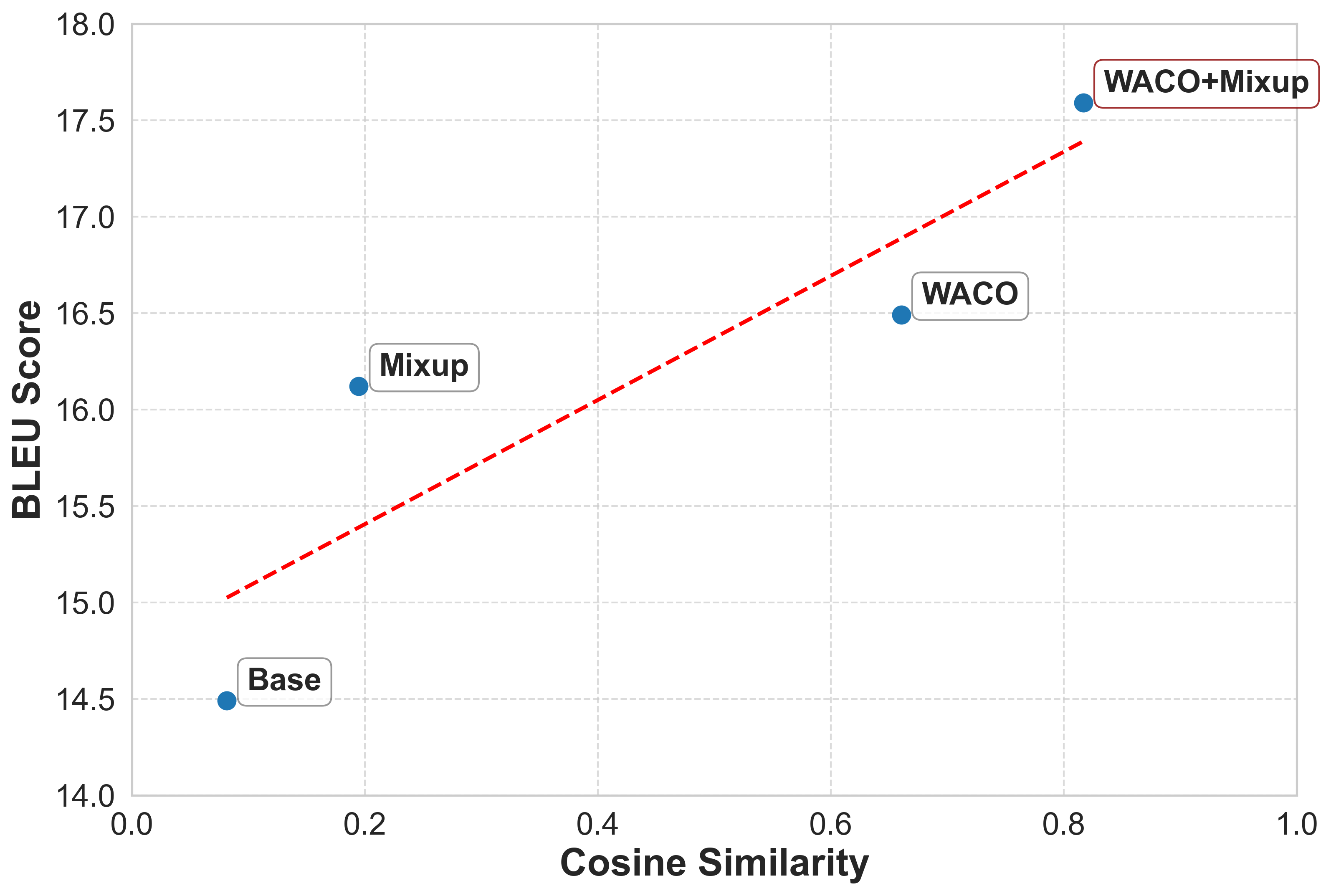}
    \caption{The figure shows the correlation between the cosine similarity of speech and text representations on En-De dev set, and the BLEU score on the adversarial test set. We see a correlation between speech and text alignment and CMRT-FN's effectiveness.}
    \label{fig:cmrt_space}
\end{figure}

Figure \ref{fig:cmrt_space} shows that the cosine similarity between speech and text representations strongly correlates with performance on adversarial data after fine-tuning with CMRT. Furthermore, although mixup training alone falls significantly behind WACO in terms of semantic similarity, the difference in BLEU score between the two is not as significant. This suggests that mixup offers a clear advantage: it familiarizes the translation encoder-decoder to mixup (speech-text) inputs, which is crucial for CMRT during the fine-tuning stage. Critically, the figure also shows that our method of combining WACO and Mixup training yields the strongest semantic alignment and the highest robustness to adversarial inputs. This confirms our hypothesis that a stronger alignment between speech and text representations should allow robustness to transfer from text to speech, and that mixup training is necessary to familiarize the model on mixup inputs.

\subsection{Effect of $\lambda_{kl}$ on CMRT-FN}
\begin{figure}[ht]
    \centering
    \includegraphics[width=0.48\textwidth]{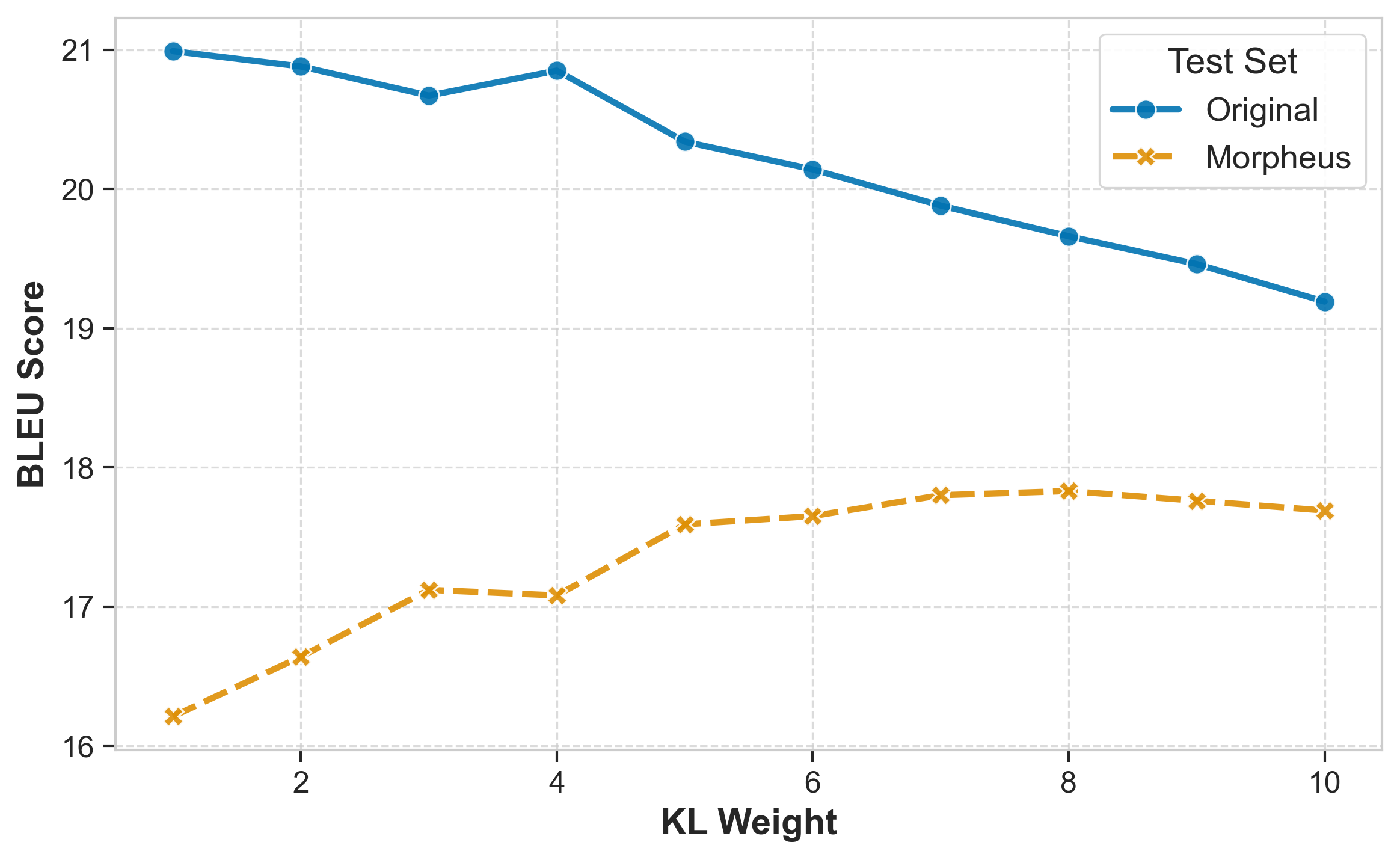}
    \caption{Effect of KL divergence weight $\lambda_{kl}$ on adversarial Morpheus and original CoVoST 2 test sets.}
    \label{fig:kl_weight}
\end{figure}

When fine-tuning the model with CMRT-FN, KL divergence serves as a regularization mechanism that penalizes the model for being sensitive to the adversarial input. As formulated in \S\ref{sec:cmrt_fn}, this is done by minimizing asymmetric KL between the output probability distribution of adversarial mixup with both speech and text output probability distributions. The weight $\lambda_{kl}$ is used to control the strength of KL as shown in Equation \ref{eq:cmrt_fn}. Figure \ref{fig:kl_weight} shows the effect of increasing the value of $\lambda_{kl}$ between $1$ and $10$ on the robustness to adversarial examples, and its effect on original CoVoST 2 test set performance. The figure shows that there is a trade-off between improved performance on Morpheus adversarial subset and performance on the original test set as we increase $\lambda_{kl}$. This is a documented trade-off between adversarial robustness and accuracy \cite{tsipras2018robustness}. Furthermore, increasing $\lambda_{kl}$ indirectly down-weights the ST loss which means the model is learning less from the original training samples. The figure also shows that adversarial robustness plateaus after $\lambda_{kl}=8$, suggesting that the method exhausted the amount of adversarial speech robustness it can learn from text. The fact that CMRT-FN plateaus below TTS-Morpheus-FN suggests that the model retains modality specific information that limits cross-modal transfer. We attribute this to the limitations of contrastive learning \cite{liang2022mind}, and will research other techniques for minimizing modality specific information in future work.
\subsection{Layer-wise Similarity Analysis}
\label{sec:layer_cka}

\begin{figure}[ht]
    \centering
    \includegraphics[width=0.48\textwidth]{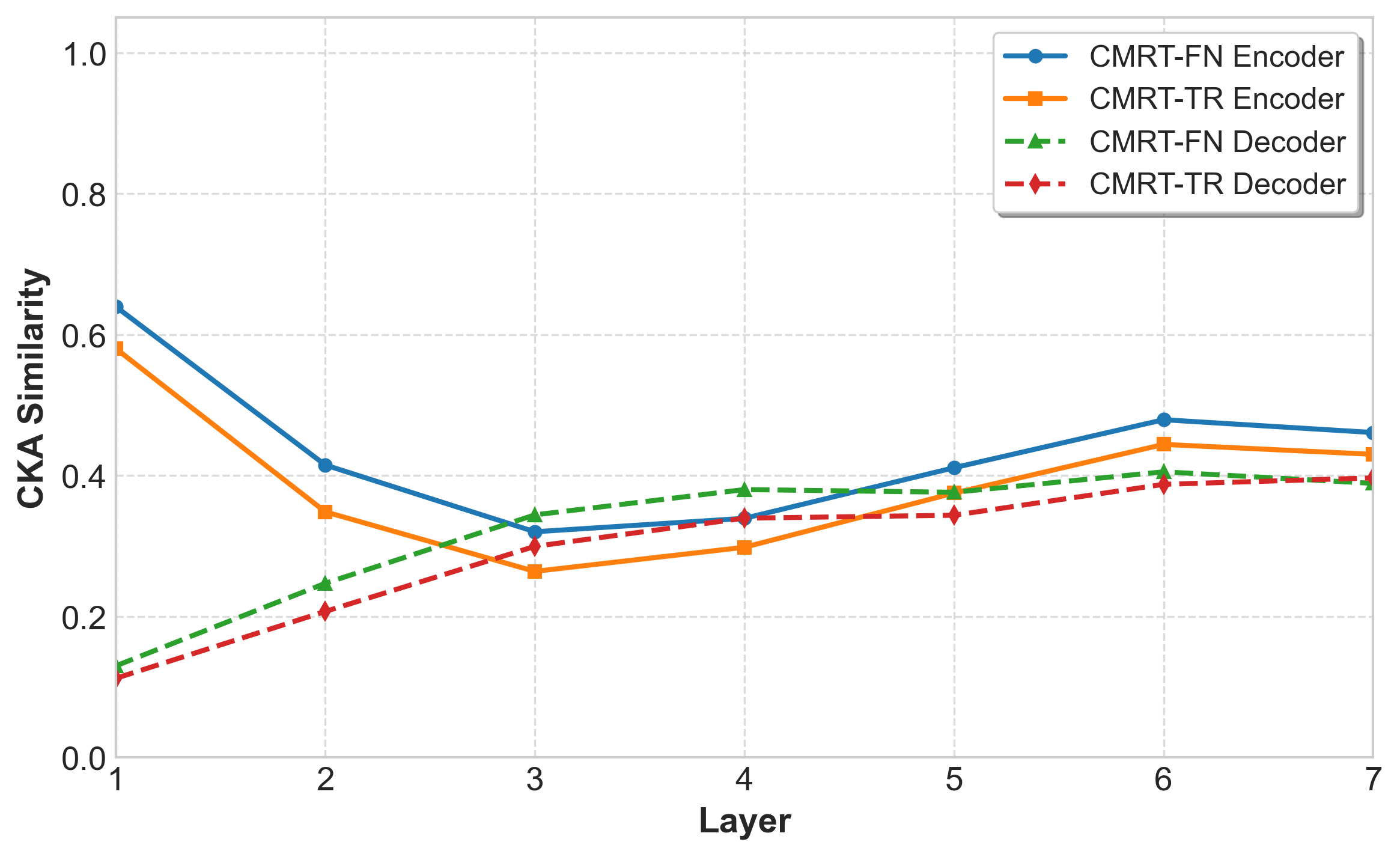}
    \caption{We compare the CKA similarity of representations of adversarial sentences of CMRT-TR and CMRT-FN models against TTS-Morpheus-FN. CMRT-FN model is more aligned with TTS-Morpheus-FN because both are trained to handle Morpheus errors.}
    \label{fig:cmrt_cka}
\end{figure}

CMRT-FN behaves similarly to fine-tuning which is robustness to inflectional morphology variation, but are there similarities on the representation level as well? To answer this, we measure how the model trained with CMRT represents adversarial speech examples compared to the model that is fine-tuned on adversarial speech. We take the mean of the sequence representations of each sentence and use Centered Kernel Alignment (CKA) \cite{kornblith0LH19} to compute the similarity of CMRT-FN and CMRT-TR with TTS-Morpheus-FN on En-De dev set. Although representations are averaged over sentences that contain both clean and adversarial words, the similarity of CMRT-FN is still  higher than CMRT-TR. This  suggests that CMRT-FN not only matches the behavior of adversarially fine-tuned models but also converges towards similar internal representations, which confirms a main claim of CMRT: robustness can be transferred from text to speech modality, even in the absence of adversarial speech data. 

\section{Conclusion}
In this work, we addressed the vulnerability of End-to-End Speech Translation models to morphological variations, a critical yet overlooked aspect of real-world robustness. We introduced Speech-MORPHEUS, a framework for evaluating speech models against inflectional perturbations, and proposed Cross-Modal Robustness Transfer (CMRT) to mitigate these vulnerabilities. Our results across four language pairs demonstrate that CMRT improves adversarial robustness by over 3 BLEU points without the computational burden of synthetic speech generation. These findings provide strong evidence that adversarial robustness can be effectively transferred across modalities through shared latent spaces. Future research will explore the extensibility of CMRT to broader speech-to-text tasks and architectures.

\section{Limitations}
Our method has two main limitations:

First, it is composed of two stages (i.e. CMRT-TR and CMRT-FN) which can increase the complexity of E2E-ST training, therefore, in future work we will aim at training for both ST and robustness in one step.

Secondly, our first stage of training where we focus on maximizing the alignment between speech and text representations leads to worse results on CoVoST 2 test set than HuBERT-Transformer baseline on En-X directions. We will investigate this negative effect of the alignment and explore ways to eliminate it.

\section*{Acknowledgments}
The research presented in this paper was conducted as part of VOXReality project\footnote{\texttt{\url{https://voxreality.eu/}}}, which was funded by the European Union Horizon Europe program under grant agreement No 101070521.

This work used the Dutch national e-infrastructure with the support of the SURF Cooperative using grant no. EINF-11297.

\bibliography{acl_latex}

\appendix

\section{Experimental Setup Details}
\label{sec:exp_setup_details}
\noindent
\textbf{Speech-Morpheus}
Morpheus was introduced initially for English only, although it was extended to other languages in \cite{jayanthi-pratapa-2021-study}, we found the French inflections to be limited to verbs. Therefore, we extend Morpheus to French by using inflecteur \footnote{\scriptsize\texttt{\href{https://github.com/Achuttarsing/inflecteur}{https://github.com/Achuttarsing/inflecteur}}}
 to find inflections for French, while we use LemmInflect \footnote{\scriptsize\texttt{\href{https://github.com/bjascob/LemmInflect}{https://github.com/bjascob/LemmInflect}}} for English similar to the original work. For tokenization and POS tagging, we use Spacy \footnote{\scriptsize\texttt{\href{https://github.com/explosion/spaCy}{https://github.com/explosion/spaCy}}} and Spacy-lefff \footnote{\scriptsize\texttt{\href{https://github.com/sammous/spacy-lefff}{https://github.com/sammous/spacy-lefff}}} for English and French respectively. Furthermore, Speech-Morpheus requires two extra tools for phonemization and TTS. For phonemization, we use eSpeak NG \footnote{\scriptsize\texttt{\href{https://github.com/espeak-ng/espeak-ng}{https://github.com/espeak-ng/espeak-ng}}}, while for TTS, we use XTTS-v2 \footnote{\scriptsize\texttt{\href{https://huggingface.co/coqui/XTTS-v2}{https://huggingface.co/coqui/XTTS-v2}}} as a state-of-the-art multilingual TTS model \cite{casanova2024-xtts}.

 \noindent
\textbf{Training}
Our models are trained in two stages, first we train the translation encoder-decoder on transcription-translation pairs from the CoVoST 2 train set. We train with a learning of 1e-4, a maximum of 33k tokens per batch, for a maximum of 100k steps, where we early stop the training if the loss doesn't decrease for 10 epochs on the dev set. In the second stage, we fine-tune the speech encoder and translation encoder-decoder for ST with a learning rate of 1e-4, a maximum of 16M audio frames, for 40k steps. For CMRT-TR, we train with a contrastive loss weight $\lambda_{ctr}=1.0$, contrastive temperature $\tau=0.2$, a KL weight $\lambda_{kl}=2.0$, and a mixup probability $p^*=0.8$ \footnote{\scriptsize{We found that decreasing the mixup probability hurts the performance on the original CoVoST 2 test set without improving speech-text alignment}}.

\noindent
\textbf{Fine-tuning}
We fine-tune models from the training stage with a learning rate of 1e-4, a maximum of 16M audio frames, for 5k steps. In this stage, the speech encoder is kept frozen while we train the rest of the model. For CMRT-FN, we increase the KL weight to $5.0$, and use the same mixup probability $p^*=0.8$.

The MT models are trained using one A100 GPU and ST models are trained using one H100 GPU. We use Fairseq \footnote{\scriptsize\texttt{\href{https://github.com/facebookresearch/fairseq}{https://github.com/facebookresearch/fairseq}}} \cite{ott-etal-2019-fairseq} for the implementation.

\noindent
\textbf{Inference}
For models from the training stage, we average the last 10 epoch checkpoints and evaluate, while for the fine-tuning stage, we take the average when the number of epochs exceeds 10 otherwise we evaluate the last checkpoint. We generate with a beam size of 5 and use SacreBLEU \cite{post-2018-call} to compute detokenized case-sensitive BLEU score \cite{papineni2002}. 

\section{CMRT Improves Robustness of Massively Pre-trained Models}
\label{sec:cmrt_nllb_results}
Table \ref{tab:cmrt_nllb_results} shows the results of using No Language Left Behind (NLLB) \cite{nllbteam2022language} as a massively pre-trained multilingual model for initializing the MT transformer. We specifically use the efficient distilled version \footnote{\scriptsize{\texttt{\href{https://huggingface.co/facebook/nllb-200-distilled-600M221}{https://huggingface.co/facebook/nllb-200-distilled-600M}}}}. The results demonstrate that although using a massively pre-trained model significantly improves the results, we still see a significant decrease in performance when evaluating on Morpheus (a drop of 5.2 from Original test set on average). This shows that the issue of robustness is inherent in training on clean data, and that fine-tuning on adversarial data can be crucial for improving robustness. More importantly, using a pre-trained model does not hinder the the robustness benefits of fine-tuning using CMRT but rather the amplifies them, where fine-tuning with CMRT leads to 4.4 and 3.3 BLEU points improvement on average over HuBERT-NLLB and CMRT-TR respectively. Furthermore, on average, CMRT leads to a minimal drop in performance on the Original test set (e.g. 0.2 and 0.3 compared to HuBERT-NLLB and CMRT-TR respectively).
\begin{table*}[ht]
    \centering
    \scriptsize
    \setlength{\tabcolsep}{4pt}
    \begin{tabular}{llccccccccccc}
        \toprule
        & \multicolumn{3}{c}{\textbf{En-De}} & \multicolumn{3}{c}{\textbf{En-Ca}} & \multicolumn{3}{c}{\textbf{En-Ar}} & \multicolumn{3}{c}{\textbf{Fr-En}} \\
        \cmidrule(lr){2-4} \cmidrule(lr){5-7} \cmidrule(lr){8-10} \cmidrule(lr){11-13}
         & Original & Morpheus & Clean & Original & Morpheus & Clean & Original & Morpheus & Clean & Original & Morpheus & Clean \\ 
        \midrule
        HuBERT-NLLB & \textbf{28.1} & 22.9 & \textbf{32.8} & 30.8 & 25.5 & 37.5 & \textbf{20.6} & 17.4 & 24.1 & 31.9 & 24.7 & 31.4 \\
        CMRT-TR & 27.6 & 23.4 & \textbf{32.8} & \textbf{31.0} & 26.1 & \textbf{38.1} & 19.6 & 17.9 & \textbf{24.4} & \textbf{33.7} & 27.2 & \textbf{34.1} \\
        TTS-Morpheus-FN (50K) & 24.9 & \textbf{29.1} & 30.3 & 27.9 & \textbf{33.3} & 35.0 & 18.4 & \textbf{21.8} & 22.5 & 30.4 & \textbf{31.5} & 31.6 \\
        CMRT-FN & 26.6 & \underline{26.5} & 32.6 & 30.9 & \underline{30.6} & \textbf{38.1} & 19.5 & \underline{20.0} & 24.1 & 33.6 & \underline{30.8} & 33.8 \\
        \bottomrule
    \end{tabular}
    \caption{Results of using NLLB pre-trained model for initializing the MT transformer. We underline the model scores that achieved the best result on Morpheus without using adversarial speech. On overage, CMRT-FN improves robustness to Speech-Morpheus adversarial attack by more than 4 BLEU points compared to HuBERT-NLLB without using any adversarial speech data.}
    \label{tab:cmrt_nllb_results}
\end{table*}

\end{document}